# Honey Adulteration Detection using Hyperspectral Imaging and Machine Learning

Mokhtar A. Al-Awadhi
*Department of Computer Science and IT*
*Dr. Babasaheb Ambedkar Marathwada University*
Aurangabad, India
mokhtar.awadhi@gmail.com

Ratnadeep R. Deshmukh
*Department of Computer Science and IT*
*Dr. Babasaheb Ambedkar Marathwada University*
Aurangabad, India
rrdeshmukh.csit@bamu.ac.in

*Abstract*— This paper aims to develop a machine learning-based system for automatically detecting honey adulteration with sugar syrup, based on honey hyperspectral imaging data. First, the floral source of a honey sample is classified by a botanical origin identification subsystem. Then, the sugar syrup adulteration is identified, and its concentration is quantified by an adulteration detection subsystem. Both subsystems consist of two steps. The first step involves extracting relevant features from the honey sample using Linear Discriminant Analysis (LDA). In the second step, we utilize the K-Nearest Neighbors (KNN) model to classify the honey botanical origin in the first subsystem and identify the adulteration level in the second subsystem. We assess the proposed system performance on a public honey hyperspectral image dataset. The result indicates that the proposed system can detect adulteration in honey with an overall cross-validation accuracy of 96.39%, making it an appropriate alternative to the current chemical-based detection methods.

I. INTRODUCTION

Food Adulteration is a common fraud practice aiming to gain fast profits by adding inexpensive substances into food, which lowers its quality and results in economic and health consequences [1]. Honey is among the liquid food that is vulnerable to adulteration due to its high economical price. Besides that, the availability of cheap industrial sugar syrups that can be added to honey without making noticeable changes in the color or taste has made honey attractive to adulteration. Honey can be adulterated directly by adding artificial sugar syrups or water into it or indirectly through overfeeding honeybees with artificial sugar. The most common industrial sugars that are used to adulterate honey include sucrose syrup, fructose syrup, glucose syrup, invert sugar syrup, high fructose corn syrup, and rice syrup.

For honey adulteration detection, established analytical approaches such as carbon isotope [2], chromatography [3], and physicochemical parameter analysis [4] have been used. Numerous studies have demonstrated the effectiveness of these procedures in detecting adulteration. However, they are costly, time-consuming, destructive, and require training and sample preparation. As a result, quick, nondestructive, and precise analytical methods are required to supplement existing techniques. Nuclear Magnetic Resonance (NMR) spectroscopy has been formerly applied to honey adulteration detection [5]. Nevertheless, this technique is time-consuming, costly, and requires sample preparation. Terahertz spectroscopy, in conjunction with multivariate analysis, has been employed to identify honey adulteration [6]. However, the detection model was not very accurate. Near-Infrared (NIR) spectroscopy [7] and Visible NIR (VIS-NIR) spectroscopy [8] obtained good results in honey adulteration detection. Compared to the traditional methods, these spectroscopic techniques are faster, nondestructive, inexpensive, and require neither training nor sample preparation.

Hyperspectral Imaging (HSI) technology extends the VIS-NIR spectroscopy to the spatial domain, where it achieves the advantages of both spectroscopic and spatial technologies. A hyperspectral imager produces a three-dimensional image. The first two dimensions represent the spatial data, while the third dimension represents the spectral data. The spectral range of a hyperspectral camera is usually from 400 to 1000 nm, covering the visible and near-infrared regions in the electromagnetic spectrum. HSI combined with Machine Learning (ML) algorithms like Support Vector Machine (SVM) and KNN achieved excellent results in classifying honey botanical origins [9]. For adulteration detection using HSI, only one study was published [10]. The study used various ML models, such as Artificial Neural Network (ANN), SVM, LDA, Fisher, and Parzen, to detect adulteration with fructose-glucose syrup at concentrations 10%, 20%, 30%, and 50%. A dataset consisting of hyperspectral images of honey samples was used to assess the performance of the models. The dataset comprised data of 56 pure and adulterated honey samples. Experimental results showed that the models achieved classification accuracies of 95%, 92%, 90%, 89%, and 84% using ANN, SVM, LDA, Fisher, and Parzen, respectively.

The various analytical methods developed in previous studies aimed at either classifying honey botanical origins [11] or detecting adulteration in honey [12]. In the present study, we propose an ML-based method for classifying honey botanical origins and detecting adulteration in honey. The data set used in this study comprises spectral instances resulting from segmenting hyperspectral images of pure and adulterated honey samples from 11 different botanical origins [13]. Figure 1 displays the spectral information of some pure and adulterated honey samples from the Manuka botanical origin. The adulterated honey samples were prepared by adding sugar syrup into pure honey samples at concentrations of 5%, 10%, 25%, and 50%. The dataset consists of 8675 rows representing spectral instances of honey samples from different floral sources. Each instance in the dataset has 128 attributes representing spectral bands from 400 nm to 1000 nm, with a spectral increment of 5 nm.

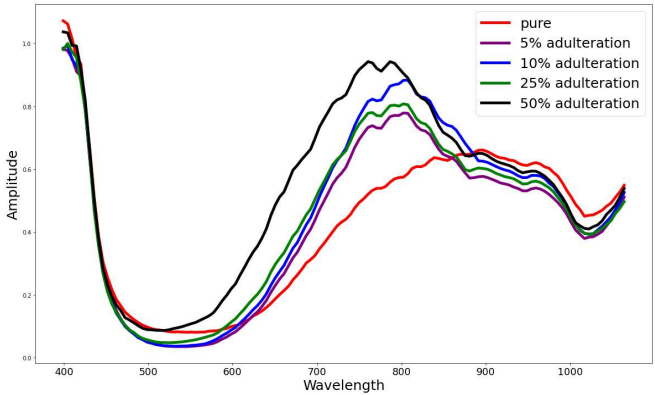

Fig. 1. The spectrum of some pure and adulterated Manuka honey samples

## II. PROPOSED SYSTEM

The honey adulteration detection system proposed in the present study consists of two subsystems: a botanical origin identification subsystem and an adulteration detection subsystem, as illustrated in figure 2. The first subsystem identifies the honey botanical origin. The second subsystem applies a botanical origin-specific pre-trained ML model to determine whether the honey sample is pure or adulterated and determine the adulteration concentration.

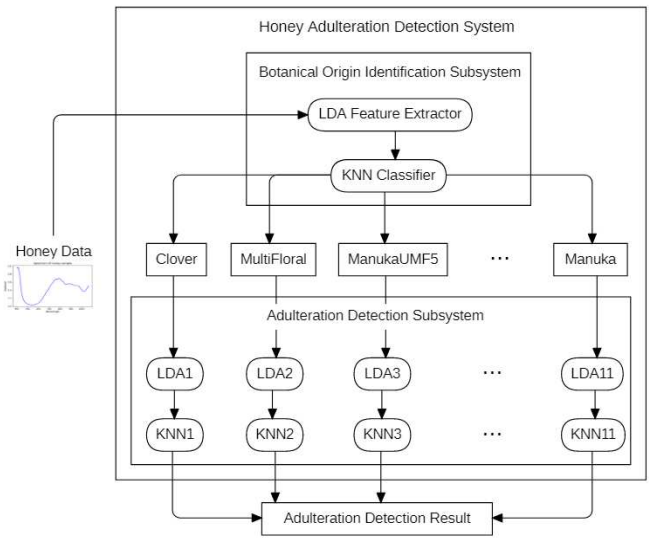

Fig. 2. Block diagram of the proposed honey adulteration detection system

### A. Botanical Origin Identification Subsystem

The function of this subsystem is to identify the botanical origin of a honey sample using its hyperspectral data and classify it into one of the floral sources present in the dataset. The identification process of the botanical origin goes over two steps: feature extraction and classification.

The feature extraction step involves extracting relevant features from the raw spectral data. Feature extraction is critical in many classification tasks because it enhances classification model performance by removing extraneous features from the dataset. In the present study, we used the LDA algorithm for feature extraction and dimensionality reduction. LDA is a supervised algorithm that can be used for feature extraction and classification. It is a linear algorithm that converts the features into a lower-dimensional space that maximizes the ratio of between-class variation to within-class variance, assuring maximal class separation [14]. We have chosen the LDA algorithm since it achieved positive results in our previous work [9] on classifying honey botanical origins. LDA is rapid, since it just involves the solution of a generic eigenvalue problem [15]. It also works for binary and multi-class tasks, and it may be extended to nonlinear LDA by using a quadratic kernel. In this research, we compare the performance of LDA with that of an unsupervised algorithm, such as the Principal Component Analysis (PCA).

In the classification step, we used the KNN model to classify the botanical origin of the honey sample. We evaluated the performance of the KNN classifier using the original features, PCA-reduced features, and LDA-reduced features. Also, we compared the performance of the KNN model to that of the SVM classifier. We have chosen the KNN and SVM classifiers since they achieved the best performance on a similar dataset in previous work [9]. The KNN classifier is a nonparametric supervised ML algorithm. It is a similarity measure-based classifier, where the similarity metric is the distance between the instances in the dataset [16]. We employed the Euclidean distance as the distance metric and a value of 5 for the parameter $k$ in the experiments, since they achieved the highest performance. The SVM algorithm is a machine learning model that finds the optimum hyperplane in a translated high-dimensional feature space to distinguish classes with the fewest errors [17]. In the experiments, we used two SVM classifiers with different kernel functions: a linear function and a Radial Basis Function (RBF).

### B. Adulteration Detection Subsystem

After the floral source of a honey sample is determined by the botanical origin identification subsystem, the adulteration detection subsystem identifies the adulteration level (sugar concentration) in the honey sample. The dataset used in the present research contains spectral data of honey samples adulterated with sugar at four concentrations: 5%, 10%, 25%, and 50%. Similar to the botanical origin identification process in the previous subsystem, the adulteration detection process goes over two steps: feature extraction using LDA and classification using KNN. There are 11 LDA models in this subsystem. Each model extracts features from honey spectral data of a particular botanical origin, where there are 11 different botanical origins in the dataset. Similarly, there is an equivalent number of KNN models. Each model classifies the adulteration levels of a particular honey type. We evaluated the performance of the KNN classifiers using the original features, PCA-reduced features, and LDA-reduced features. Also, we compared the performance of the KNN model to that of the SVM model.

### C. Performance Evaluation

To evaluate the performance of the honey adulteration detection system developed in the present study, we have used the balanced accuracy metric [18]. We have chosen the balanced accuracy since the dataset used in the current research is imbalanced. The balanced accuracy provides an accurate measure of the accuracy of each class in the dataset and can be computed using equation 1.

$$Balanced\ Accuracy = \frac{TP}{TP+FN} + \frac{TN}{TN+FP} \quad (1)$$

The True-Positive (TP) and True-Negative (TN) numbers represent the number of positively and negatively identified instances, respectively. The False-Positive (FP) and False-Negative (FN) statistics reveal the number of positive and negative instances that were wrongly recognized. We used cross-validation at 20 folds to assess the efficacy of the adulteration detection system. The training set in each fold includes examples from three acquisition numbers, while the test set contains examples from the other acquisition numbers. Further, we divide the spectral instances in the dataset such that all instances belonging to an HSI picture are included either in the training set or in the test set.

### III. RESULTS

#### A. Botanical Origin Identification

Table I displays the cross-validation accuracy and standard deviation of the classifiers. The results show the performance of the classifiers using three different feature sets: the original features, PCA-reduced features, and LDA-reduced features. Figure 3 visualizes the results listed in Table I. The results demonstrate that the proposed system achieved the best performance for classifying honey botanical origins using the LDA-reduced features and the KNN classifier, where it obtained a correct classification accuracy of 97.01%. The KNN classifier outperformed the SVM classifier on this dataset using all the feature sets, where it achieved cross-validation accuracies of 94.74%, 94.59%, and 97.01% using the original features, PCA-reduced features, and LDA-reduced features, respectively. The lowest accuracy was 82.24%, which was obtained using the RBF SVM classifier using the original features. The results also show that the PCA did not improve the performance of the classifiers, except the RBF SVM classifier. On the other hand, the LDA algorithm has significantly improved the performance of all classifiers.

TABLE I. PERFORMANCE OF THE ML MODELS FOR IDENTIFYING HONEY BOTANICAL ORIGINS USING DIFFERENT FEATURE SETS

| ML Classifier | Original features | PCA features | LDA features |
|---|---|---|---|
| KNN | 0.9474 ± 0.0223 | 0.9459 ± 0.0223 | **0.9701 ± 0.0253** |
| Linear SVM | 0.8942 ± 0.0110 | 0.8872 ± 0.0096 | 0.9672 ± 0.0215 |
| RBF SVM | 0.8224 ± 0.0136 | 0.9271 ± 0.0137 | 0.9685 ± 0.0154 |

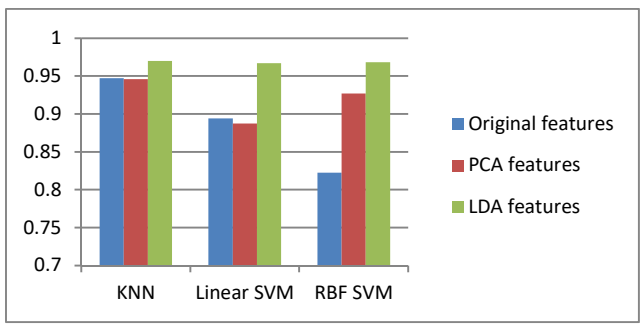

Fig. 3. The classification accuracy of the ML models for identifying honey botanical origins using different feature sets

Figure 4 shows the performance of the classifiers for a different number of LDA features. The graph shows that the classifiers achieved the best performance using the first 10 LDA features.

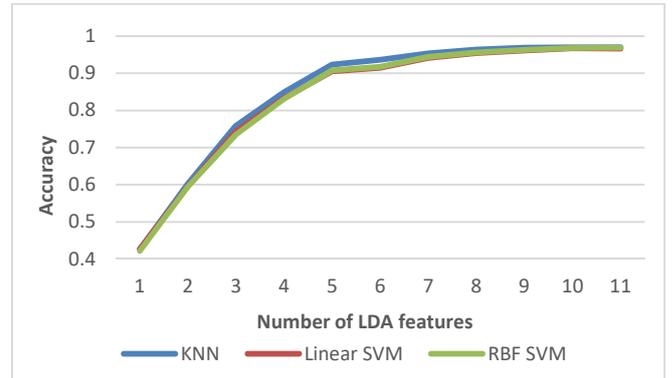

Fig. 4. Performance of the ML models using a different number of LDA features

#### B. Adulteration Detection

The performance of the KNN and SVM classifiers for detecting the adulteration in various honey types using the original features, PCA features, and LDA feature is shown in Tables II, III, and IV, respectively. Figure 5 visualizes the average cross-validation accuracy of the ML models using the three feature sets. The results in the tables show that the proposed system successfully detects adulteration in honey with an average cross-validation accuracy of 96.39% using the LDA-reduced features and the KNN classifier. The adulteration detection performance of the classifiers varied according to the type of the honey, where the adulteration in some honey types, such as ManukaUMF5, ManukaUMF15, and ManukaUMF10, was accurately detected.

Results also show that feature extraction using PCA has improved the performance of the RBF SVM classifier only. On the other hand, the performance of all the classifiers has been significantly improved by dimensionality reduction using LDA. Findings in Tables II and IV show that KNN outperforms SVM in detecting sugar adulteration using the original features and LDA features, where it achieved accuracies of 91.16% and 96.39% using the original features and LDA features, respectively. On the other hand, RBF SVM outperformed KNN using PCA features, where it obtained an accuracy of 91.03%.

TABLE II. PERFORMANCE OF THE ML MODELS FOR DETECTING THE ADULTERATION IN DIFFERENT HONEY TYPES USING ORIGINAL FEATURES

| Botanical Origin | KNN | Linear SVM | RBF SVM |
|---|---|---|---|
| Clover | 0.876 ± 0.0739 | 0.8583 ± 0.0733 | 0.7146 ± 0.0913 |
| Multifloral | 0.9547 ± 0.0711 | 0.9236 ± 0.0717 | 0.9069 ± 0.0632 |
| ManukaUMF5 | 01.00 ± 0.0000 | 01.00 ± 0.0000 | 01.00 ± 0.0000 |
| ManukaUMF15 | 0.9832 ± 0.0348 | 0.9708 ± 0.0312 | 0.9055 ± 0.0197 |
| ManukaUMF20 | 0.6787 ± 0.0951 | 0.7723 ± 0.0910 | 0.7580 ± 0.0793 |
| ManukaUMF10 | 0.9595 ± 0.0387 | 0.9773 ± 0.0337 | 0.9182 ± 0.0604 |
| ManukaBlend | 0.9144 ± 0.0552 | 0.9207 ± 0.0436 | 0.8720 ± 0.0572 |
| BorageField | 0.9511 ± 0.0567 | 0.9476 ± 0.0567 | 0.8971 ± 0.0608 |
| Kamahi | 0.8069 ± 0.0765 | 0.8040 ± 0.0727 | 0.7533 ± 0.0830 |
| Rewarewa | 0.9251 ± 0.0688 | 0.8482 ± 0.0823 | 0.7553 ± 0.0821 |
| Manuka | 0.9779 ± 0.0208 | 0.9861 ± 0.0077 | 0.9557 ± 0.0152 |
| Average | 0.9116 ± 0.054 | 0.9099 ± 0.051 | 0.8579 ± 0.056 |

TABLE III. PERFORMANCE OF THE ML MODELS FOR DETECTING THE ADULTERATION IN DIFFERENT HONEY TYPES USING PCA FEATURES

| Botanical Origin | KNN | Linear SVM | RBF SVM |
|---|---|---|---|
| Clover | 0.8733 ± 0.0750 | 0.7942 ± 0.0562 | 0.8512 ± 0.0694 |
| Multifloral | 0.9289 ± 0.0640 | 0.9127 ± 0.0663 | 0.9160 ± 0.0726 |
| ManukaUMF5 | 1.00 ± 0.0000 | 1.00 ± 0.0000 | 1.00 ± 0.0000 |
| ManukaUMF15 | 0.9712 ± 0.0521 | 0.9210 ± 0.0349 | 0.9677 ± 0.0523 |
| ManukaUMF20 | 0.6622 ± 0.0930 | 0.7713 ± 0.0906 | 0.7222 ± 0.1169 |
| ManukaUMF10 | 0.9592 ± 0.0404 | 0.9760 ± 0.0336 | 0.9765 ± 0.0376 |
| ManukaBlend | 0.9131 ± 0.0581 | 0.9152 ± 0.0468 | 0.9433 ± 0.0531 |
| BorageField | 0.9505 ± 0.0589 | 0.9447 ± 0.0561 | 0.9777 ± 0.0462 |
| Kamahi | 0.7979 ± 0.0711 | 0.7880 ± 0.0766 | 0.8589 ± 0.0887 |
| Rewarewa | 0.8202 ± 0.0833 | 0.7933 ± 0.0767 | 0.8080 ± 0.1184 |
| Manuka | 0.9759 ± 0.0221 | 0.9801 ± 0.0095 | 0.9914 ± 0.0075 |
| Average | 0.8957 ± 0.056 | 0.8906 ± 0.050 | 0.9103 ± 0.060 |

TABLE IV. PERFORMANCE OF THE ML MODELS FOR DETECTING THE ADULTERATION IN DIFFERENT HONEY TYPES USING LDA FEATURES

| Botanical Origin | KNN | Linear SVM | RBF SVM |
|---|---|---|---|
| Clover | 0.9052 ± 0.0726 | 0.9031 ± 0.0756 | 0.9081 ± 0.0752 |
| Multifloral | 0.9278 ± 0.0865 | 0.9293 ± 0.0873 | 0.9289 ± 0.0869 |
| ManukaUMF5 | 1.00 ± 0.0000 | 1.00 ± 0.0000 | 1.00 ± 0.0000 |
| ManukaUMF15 | 1.00 ± 0.0000 | 1.00 ± 0.0000 | 1.00 ± 0.0000 |
| ManukaUMF20 | 0.9765 ± 0.0401 | 0.9757 ± 0.0426 | 0.9627 ± 0.0451 |
| ManukaUMF10 | 1.00 ± 0.0000 | 1.00 ± 0.0000 | 1.00 ± 0.0000 |
| ManukaBlend | 0.9243 ± 0.0473 | 0.9240 ± 0.0454 | 0.9211 ± 0.0470 |
| BorageField | 0.9876 ± 0.0291 | 0.9909 ± 0.0199 | 0.9881 ± 0.0277 |
| Kamahi | 0.9037 ± 0.0661 | 0.9009 ± 0.0660 | 0.9043 ± 0.0665 |
| Rewarewa | 0.9849 ± 0.0317 | 0.9847 ± 0.0319 | 0.9851 ± 0.0317 |
| Manuka | 0.9933 ± 0.0153 | 0.9913 ± 0.0184 | 0.9937 ± 0.0151 |
| Average | **0.9639 ± 0.035** | 0.9636 ± 0.035 | 0.9629 ± 0.036 |

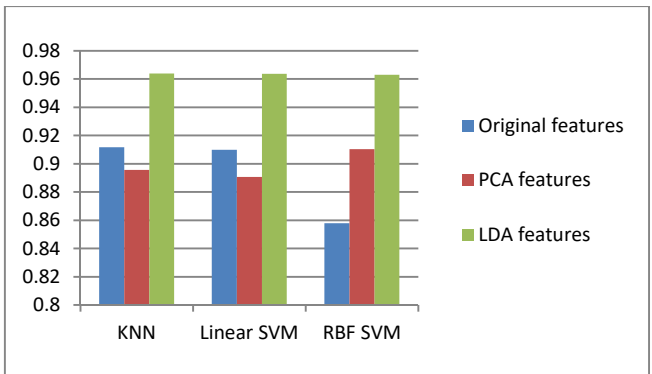

Fig. 5. Performance of the ML models for detecting the adulteration using different feature sets

Figure 6 depicts the graphical user interface of the adulteration detection system created in this study. The system receives hyperspectral data of a honey sample and displays it on the user interface. The botanical origin identification subsystem detects the floral source of the honey sample when a user clicks on the Test Sample button. The adulteration detection subsystem next assesses whether the honey sample is pure or adulterated. The user interface reveals the botanical origin as well as the amount of adulteration. A zero percent adulteration level indicates that the honey sample is pure. The picture depicts a plot of spectral data from an example honey sample, as well as the recognized botanical origin and amount of adulteration.

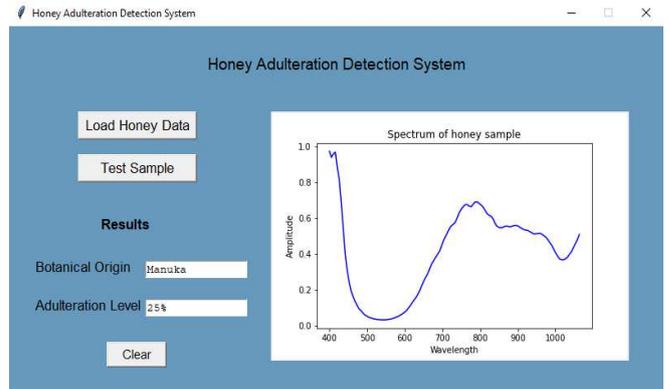

Fig. 6. The graphical user interface of the adulteration detection system

## IV. DISCUSSION

In the present paper, we developed an ML-based method for classifying honey botanical origins and detecting adulteration in honey. Compared to previous work on HSI honey data, the system developed in the current study has achieved the highest accuracy, utilizing the LDA and KNN algorithms. The developed system provides fast, automatic, and nondestructive detection of honey adulteration using honey hyperspectral imaging data. As seen in the last column of Table I, the KNN and SVM classifiers performed well in distinguishing between the diverse honey botanical sources. The results in the table confirm the discriminative power of the LDA algorithm, where the performance of all classifiers was significantly improved using the LDA reduced features. The LDA algorithm outperformed the PCA algorithm since the classes using LDA become more separable and the distances between instances within the same class were minimal, as illustrated in figures 7 and 8. All KNN and SVM models achieved excellent average classification accuracy with small differences, as shown in Table IV. The standard deviations of the classifiers were tiny, showing stable performance. The models could not achieve full discrimination of the honey types because instances of some classes in the dataset were overlapped, as illustrated in figure 8.

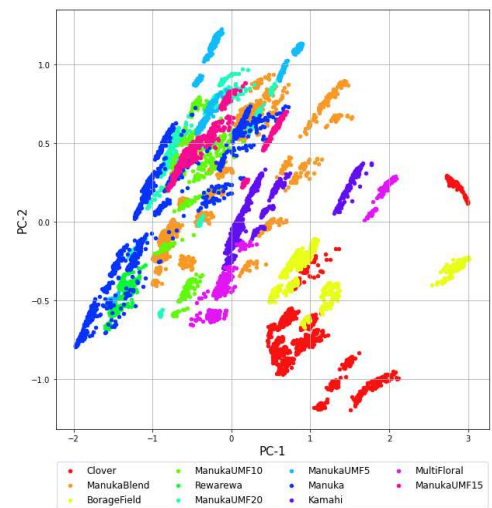

Fig. 7. Mapping of the honey instances on the first and second principal components

The performance of the KNN and SVM models in detecting honey adulteration has been improved using the LDA reduced features. Tables II, III, IV show that classifiers' performance varied depending on the honey botanical source. The sugar adulterant concentrations were not equal for all honey types in the dataset. Consequently, the number of class labels of each honey type was different; hence, the performance of the classifiers varied.

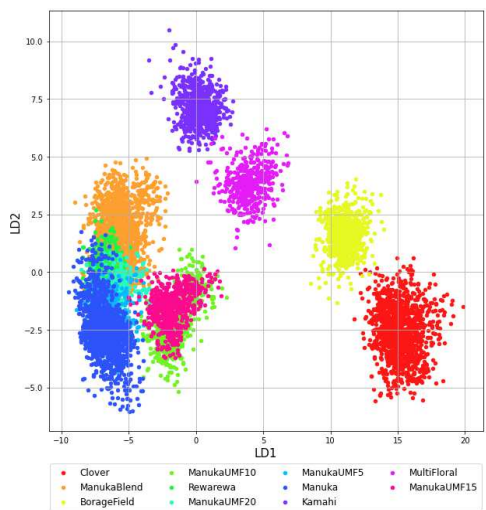

Fig. 8. Mapping of the honey instances on the first and second linear discriminant components

Findings in Table IV demonstrate that the classifiers accurately detected sugar adulteration in most honey types in the dataset such as the Manuka honey with the unique factors 5, 15, and 10. Table V shows a summary of the proposed system and past research on another HSI dataset. Results confirm the effectiveness of HSI and ML for detecting adulteration in honey. The performance of the classifiers varied due to the differences in the adulterants types, honey floral sources, and adulteration concentrations.

TABLE V.  SUMMARY OF THE PROPOSED SYSTEM AND PREVIOUS WORK ON ANOTHER HSI DATASET

| Reference | Adulterant | Adulteration Levels | Honey Types | Model | Accuracy |
|---|---|---|---|---|---|
| S. Shafiee et al. [10] | Fructose-glucose syrup | 10%, 20%, 30%, 50% | 7 | ANN<br>SVM | 95%<br>92% |
| Proposed System | Sugar syrup | 5%, 10%, 25%, 50% | 11 | KNN<br>SVM | 96.39%<br>96.36% |

## V. CONCLUSIONS AND FUTURE WORK

The current study developed a system for automated and nondestructive detection of sugar adulteration in honey utilizing hyperspectral imaging and machine learning. The developed adulteration detection system first identifies the botanical origin of a honey sample. Then, the honey sample is tested for sugar syrup adulteration. The proposed system's performance was evaluated using a publically accessible dataset that contained hyperspectral imaging data of pure and contaminated honey from various botanical sources. The developed system successfully detected adulteration in honey with an overall cross-validation accuracy of 96.39%. The findings demonstrate the capabilities of hyperspectral imaging in conjunction with ML for providing a rapid, low-cost, and noninvasive tool for detecting honey adulteration. Our future work will use regression models to improve the performance of our system.